\documentclass[10pt,twocolumn,letterpaper]{article}
\pdfoutput=1
\usepackage{cvpr}
\usepackage{times}
\usepackage{epsfig}
\usepackage{graphicx}
\usepackage{amsmath}
\usepackage{amssymb}
\usepackage[utf8]{inputenc}

\usepackage[breaklinks=true,bookmarks=false]{hyperref}

\cvprfinalcopy 

\def\cvprPaperID{3798} 

\ifcvprfinal\fi
\begin{document}

\title{Convolutional Relational Machine for Group Activity Recognition}

\author{Sina Mokhtarzadeh Azar${^1}$\thanks{Equal contribution} , Mina Ghadimi Atigh${^1}$\footnotemark[1] , Ahmad Nickabadi${^1}$, Alexandre Alahi${^2}$\\
${^1}$ Amirkabir University of Technology (AUT), SML lab\\
424 Hafez Ave, Tehran, Iran\\
{\tt\small \{sinamokhtarzadeh,minaghadimi\}@aut.ac.ir}
\and
${^2}$ École Polytechnique Fédérale de Lausanne (EPFL), VITA lab\\
CH-1015 Lausanne, Switzerland\\
}

\maketitle
\thispagestyle{empty}

\begin{abstract}
  We present an end-to-end deep Convolutional Neural Network called Convolutional Relational Machine (CRM) for recognizing group activities that utilizes the information in spatial relations between individual persons in image or video. It learns to produce an intermediate spatial representation (activity map) based on individual and group activities. A multi-stage refinement component is responsible for decreasing the incorrect predictions in the activity map. 
  Finally, an aggregation component uses the refined information to recognize group activities. Experimental results demonstrate the constructive contribution of the information extracted and represented in the form of the activity map. CRM shows advantages over state-of-the-art models on Volleyball and Collective Activity datasets.

\end{abstract}

\section{Introduction}
    Human activities can be categorized into two types: either individual actions involving a single person \eg \textit{Running/Jumping}, or group  activities involving multiple humans \eg \textit{Talking/Queuing}. Recognizing group activity requires understanding individual actions as well as the joint modeling of the group of individuals. It is important in applications like sports video analysis, surveillance, and even social robots that need to operate in a socially-aware manner around humans (\eg not crossing two individuals talking to each other). In this work, we propose to classify the group activity given single or few consecutive images of a scene. We do not explicitly detect nor track any individual. 

\begin{figure}[t]
\begin{center}
   \includegraphics[trim=0 0 0 0, width=1.0\linewidth]{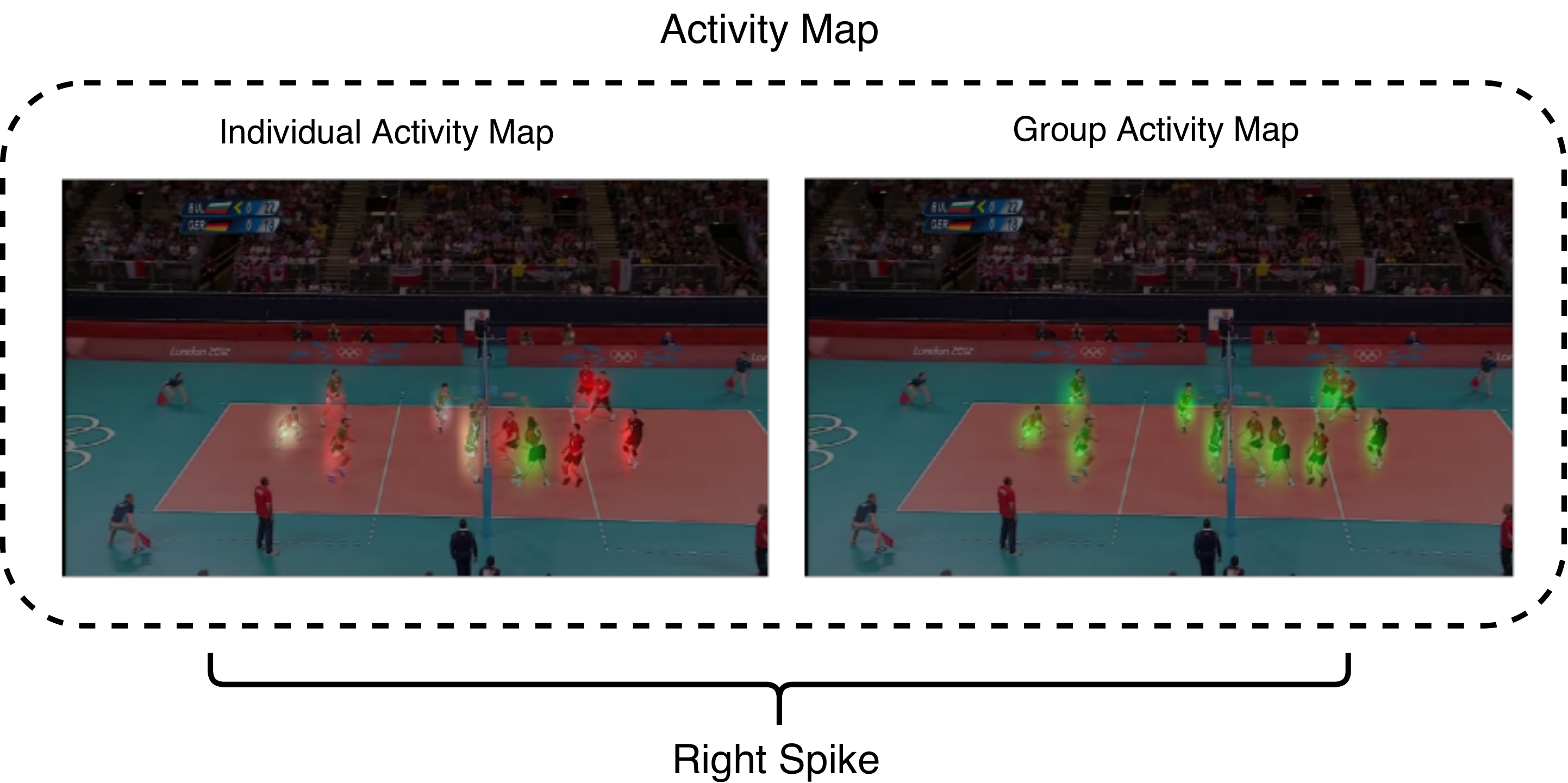}
\end{center}
   \caption{Group activity recognition using the proposed spatial representation (activity map). Our model learns to encode spatial relations in the form of the activity map and uses it to recognize the group activity (``Right Spike'' in this example). 
  Different individual and group activity fields are shown with different colors.
    } 
\label{fig:pull}
\end{figure}
There are multiple sources of information that can be helpful in recognizing activities. One source is the appearance features of individuals. Many activities are recognizable by knowing how the persons look like. Temporal information between consecutive frames of a video also plays an important role in understanding complex activities. Another recently explored source of information in reasoning over group activities is the relationships between individuals in a scene. Some of the existing approaches use various pooling strategies to extract a part of relational cues \cite{bagautdinov2017social,ramanathan2016detecting,ibrahim2016hierarchical}. In \cite{ibrahim2018hierarchical}, a more complex approach is used to extract relational information from the neighbouring individuals.
Yet, existing methods are not fully exploiting all the available information to model spatial relationships between persons. One may expect a Convolutional Neural Network (CNN) to extract these relations. Our experiments show that it is very difficult for a CNN to learn high-level relationships. Hence, we introduce a mechanism to jointly consider the relationships between individuals.

        In this paper, we propose a Convolutional Neural Network model for group activity recognition, referred to as Convolutional Relational Machine (CRM), where we introduce an intermediate activity-based representation -- an \textit{activity map}-- as a means of extracting the spatial relations between activities of persons. Along with this representation scheme, we train our architecture with a multi-stage method similar to \cite{wei2016convolutional}. Our method generates an initial activity map from the input image or video and then refines it through few steps. The refined activity map is combined with the image or video features and a final aggregation module is responsible for classifying the group activity (see Figure~\ref{fig:pull}). Our experiments show that we outperform previous works on two publicly available datasets.


\section{Related Works}

First, we present an overview of the works in action recognition. Then, some of the key works on group activity recognition are reviewed. 

\subsection{Action Recognition}
Many of the recent works in action recognition utilize a two-stream CNN model with input RGB and handcrafted optical flow features extracted from consecutive frames of a video \cite{simonyan2014two, feichtenhofer2016convolutional, wang2016temporal, carreira2017quo}. The two-stream approach was first used in \cite{simonyan2014two} by training a CNN on the single RGB frame to extract the appearance cues and another CNN on stacked optical flow fields to consider the temporal information. These streams are combined using simple fusion methods. Feichtenhofer \etal~\cite{feichtenhofer2016convolutional} study more complex fusion strategies for combining the two streams. In \cite{wang2016temporal}, Temporal Segment Network is proposed to model long-range temporal cues by dividing the video into segments and applying a multi-stream model on the snippets sampled from these segments. In a different approach, \cite{carreira2017quo} converts existing 2D CNNs to the ones with 3D filters to be able to process multiple RGB frames together. Interestingly, they still find it useful to apply their model to multiple optical flow fields and fuse the results with the RGB stream. Some other works use recurrent approaches to model the actions in video \cite{donahue2015long, Luo_2017_CVPR, Nakamura_2017_CVPR,li2018videolstm} or even a single CNN \cite{haque2017towards}. Donahue \etal~\cite{donahue2015long} propose the Long-term Recurrent Convolutional Networks model that combines the CNN features from multiple frames using an LSTM to recognize actions. In another work \cite{li2018videolstm}, VideoLSTM is developed for action classification and localization. A spatial attention mechanism is used in this model which utilizes the motion information from optical flow fields between frames.

\subsection{Group Activity Recognition}
Initial approaches for group activity recognition were based on probabilistic graphical models. \cite{lan2012discriminative} models the person-person and person-group relations in a graphical model. The optimal graph connectivity and the best set of individual actions and group activity labels are inferred in this approach. 
 A joint probabilistic framework is proposed in \cite{choi2012unified} for tracking individuals and inferring their group activity.
 
Considering the recent success of deep neural networks in the field of computer vision, various works studied the group activity recognition using deep learning. Deng \etal~\cite{deng2015deep} produces unary potentials using CNN classifiers and develops a neural network that performs message passing to refine the initial predictions. In \cite{deng2016structure}, message passing is performed in a graph with the person and group nodes by a Recurrent Neural Networks (RNN). The connections of this graph are controlled by some gating functions. 

Many of the recent deep learning based works on group activity recognition utilize the power of RNN to model the activity recognition considering the temporal domain \cite{ibrahim2016hierarchical,ramanathan2016detecting,tsunoda2017football,wang2017recurrent,shu2017cern,bagautdinov2017social,li2017sbgar,ibrahim2018hierarchical}. The idea of using RNNs for group activity recognition started with \cite{ibrahim2016hierarchical} that uses Long Short-Term Memory (LSTM) networks to model individual persons and pools the representations from them to a specific LSTM for modeling group activity. In \cite{ramanathan2016detecting}, attention pooling is utilized to give higher importance to key actors. Person-centered features are introduced in \cite{tsunoda2017football} as input to a hierarchical LSTM. Wang \etal~\cite{wang2017recurrent} introduce a three-level model based on person, group and scene representations. An LSTM models every person similar to previous models. The output representations of these LSTMS are
spatio-temporally grouped and processed to form the group representations which are then used to make scene-level predictions. 
In \cite{shu2017cern}, a new energy layer is used instead of the softmax layer which also considers the p-values of predictions. Bagautdinov \etal~\cite{bagautdinov2017social} introduce an end-to-end framework for joint detection, individual action classification and group activity recognition. In a different work, authors in \cite{li2017sbgar} develop a model that recognizes group activity based on the semantic information in the form of automatically generated captions. Most recently, Ibrahim \etal~\cite{ibrahim2018hierarchical} propose a hierarchical relational network to produce representations based on the relations of persons. Our work has a similar goal to \cite{ibrahim2018hierarchical}. 


\begin{figure*}[t]
\begin{center}
   \includegraphics[trim=0 330 20 0, width=1.0\linewidth]{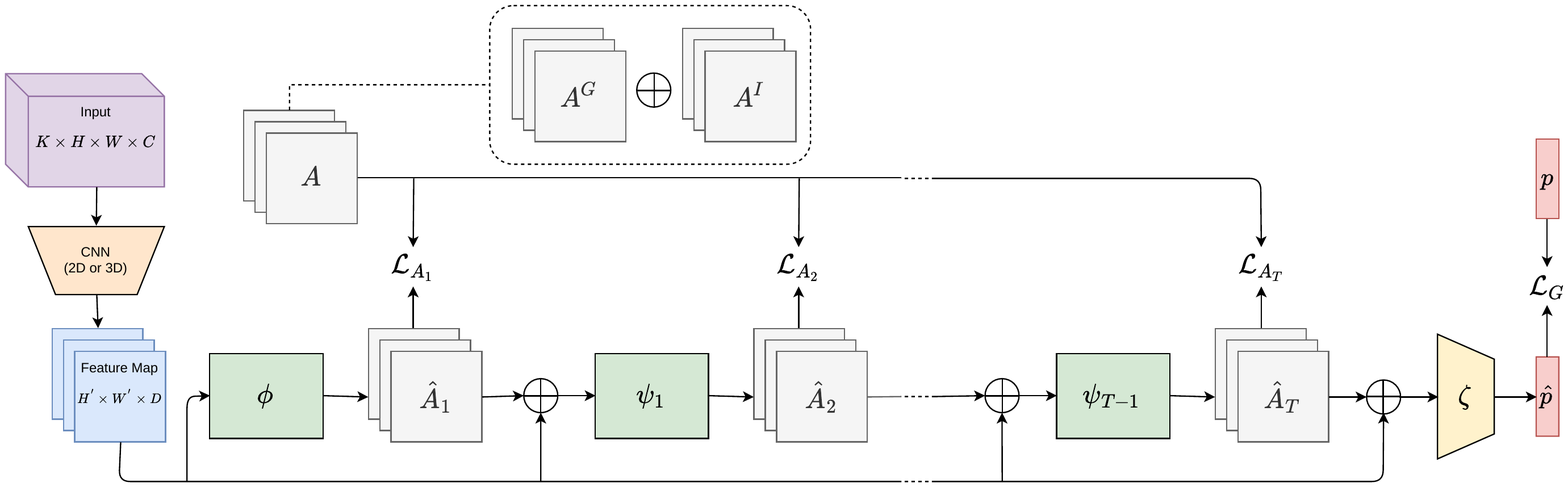}
\end{center}
   \caption{Overview of our proposed model. Input image or video is processed by a 2D or 3D CNN to produce the preliminary feature map $F$ which is then used in both activity map prediction and group activity recognition steps. The initial activity map $\hat{A}_1$ that contains individual and group activity maps in a spatial representation is generated by $\phi$. Next, the activity map is refined in multiple stages using $\psi_t$. Finally, the aggregation component $\zeta$ combines the information from $F$ and refined activity map $\hat{A}_T$ to predict the final group activity label $\hat{p}$. $\mathcal{L}_{A_t}$ is the Euclidean loss between predicted and ground truth activity maps in stage $t$ and $\mathcal{L_G}$ is the cross entropy loss for group activity.
   The ground truth activity map $A$ is composed of group activity map $A^G$ and individual activity map $A^I$.} 
\label{fig:pipeline}
\end{figure*}

\section{Proposed Model}
    The goal of our approach is to improve the group activity recognition performance by introducing an intermediate activity-based representation that we call activity map as a means of extracting the spatial relations between activities of persons. 
    Along with this representation scheme, a multi-stage neural network model is proposed for generating an initial activity map from the input image or video and then refining it in consequent steps. The final activity map is used to conduct the higher level reasoning for predicting the group activity. In the following subsection, we first review the overall outline of the model and then discuss the important components of the model in more details.

\subsection{Overview}
An overview of our model is presented in Figure~\ref{fig:pipeline}. Denote $X \in R^{K\times H \times W \times C}$ as the input to the CRM where $H$ and $W$ are the height and width of the input and $K$ and $C$ are the number of frames and channels, respectively. The input is compatible with both 3D CNNs on 4-dimensional input sequences of frames and regular CNNs on the 3-dimensional RGB single frames after squeezing the first dimension when $K=1$. A feature map is extracted by a CNN and spatially resized to $H^{'} \times W^{'}$. A 3-dimensional feature map is used in this work. Therefore, in the case of using a 3D CNN, the average of feature map from different time dimensions is calculated to form the input feature map $F \in R^{H^{'} \times W^{'} \times D}$ for the remaining parts of the model. $D$ is the number of filters in the chosen layer of the CNN. 

Given the feature map $F$ as input to the CRM, it produces the activity map $\hat{A}_t \in R^{H^{'} \times W^{'} \times N}$ in every stage $t$ with $N = N_I + N_G$ where $N_I$ and $N_G$ are the number of individual and group activity classes, respectively. These refinement stages will result in a final refined activity map $\hat{A}_T$ where $T$ is the number of stages. The corrected activity map $\hat{A}_T$ along with the feature map $F$ are given as input to the aggregation component ($\zeta$) of CRM to make a final decision for the group activity based on the relations between the activities of all the persons in the scene. 

\subsection{Activity map}

Activity map is defined as a set of 2D fields representing individual and group activities considering the bounding boxes of persons. This means there is a map for every individual and group activity in the activity map. Each map has higher values inside the bounding boxes of persons with the specific individual action or group activity label. Activity map makes it possible for the model to extract the information from spatial relations between persons. Therefore, the ground truth activity map  $A \in R^{H^{'} \times W^{'} \times N}$ is generated as a combination of individual activity map $A^I \in R^{H^{'} \times W^{'} \times N_I}$ and group activity map $A^G \in R^{H^{'} \times W^{'} \times N_G}$. 
 Particularly, a 2D Gaussian probability density function is used based on the bounding boxes of persons and their individual and group activities to construct the activity map. During the training, considering $M$ persons in the input, the activity map is created by getting the ground truth person bounding boxes $B \in R^{M \times 4}$, their individual action classes $I \in {\{1,\dots,N_I\}}^{M}$, and the group activity class $G \in \{1,\dots,N_G\}$. For each person $m$ with bounding box $B_m = (x_m^1, y_m^1, x_m^2, y_m^2)$, individual action $i$, and group activity $g$, a person specific activity map $A_m$ is produced as:
\begin{multline}
 f_m(z) = \\ \frac{1}{2\pi\sqrt{\det{\Sigma_m}}} exp\left(\frac{-1}{2}{\left(z-\mu_m\right)}^T{\Sigma_m}^{-1}\left(z-\mu_m\right)\right),
\end{multline}

\begin{equation}
 \mu_m = (\mu_m^x, \mu_m^y), \mu_m^x = \frac{x_m^1 + x_m^2}{2}, \mu_m^y = \frac{y_m^1 + y_m^2}{2} ,
\end{equation}
\begin{equation}
 \Sigma_m = \begin{bmatrix}
    {\sigma_m^x}^2 & 0 \\
    0 & {\sigma_m^y}^2 
\end{bmatrix}, \sigma_m^x = \frac{x_m^2 - x_m^1}{4}, \sigma_m^y = \frac{y_m^2 - y_m^1}{4} ,
\end{equation}
where $f_m(z)$ is calculated for every point $z$ in the fields $i$ and $N_I + g$ of the person-specific activity map $A_m$. $\mu_m$ is the middle of the bounding box for the person $m$. $\Sigma_m$ is a covariance matrix calculated for the bounding box based on the width and height of it. Bounding boxes are along the vertical and horizontal lines which makes the covariance matrices diagonal. A bivariate Gaussian map will be created on the location of the person's bounding box in the individual action and group activity fields of the person-specific activity map. Each field is separately normalized by dividing the values by the maximum value in the field. Finally, all the person-specific activity maps ($A_m$s) are aligned and combined by taking the maximum value for the aligned points to form the final activity map $A$. As a result, for a single input, persons will have 2D Gaussian maps in various individual activity fields in the activity map depending on the individual activity label of each person but in the group activity map only one of the fields will have 2D Gaussian maps for all the persons and the other group activity fields will be zero. At test time, the model will have to produce the activity map based on the input feature map $F$. It should be noted that in the cases where the segmentation masks for the persons are available, there won't be a need for the 2D Gaussian maps and the person segmentation masks can be a better choice. 

\subsection{Convolutional Relational Machine}
Convolutional Relational Machine has two major parts of refinement and aggregation. In the refinement part, similar to convolutional pose machine \cite{wei2016convolutional}, the CRM produces an initial prediction in the first stage and refines it in the next stages. After refinement, an aggregation component is responsible to aggregate the information from the whole scene and the refined activity map to produce the group activity label.  

\subsubsection{Refinement}
Given the input feature map $F$, a function $\phi$ works on it to produce the initial activity map $\hat{A}_1$ in the first stage. In the following step, another function $\psi_1$ works on the combination of $F$ and $\hat{A}_1$ to produce the refined activity map $\hat{A}_2$. The refinement component of the CRM can be written as follows:
\begin{equation}
 \begin{split}
 \hat{A}_1 & = \phi(F), \\
 \hat{A}_t & = {\psi_{t-1}}(F \oplus \hat{A}_{t-1}),  1 < t <= T,
 \end{split}
\end{equation}
where T is the number of stages and $\oplus$ shows the concatenation. $\phi$ and $\psi$s are made of a series of convolutional layers. Denote $conv(x)$ as a convolutional layer with kernel size $x$ and stride $1 $. $\phi$ is composed of three $conv(3)$ followed by two $conv(1)$.  All the $\psi_t$s have the same layers which is different from the layers in $\phi$. Three $conv(7)$ and two $conv(1)$ are used to construct the $\psi_t$.
See the supplementary material for the details of these layers.

\subsubsection{Aggregation}
Given the refined activity map $\hat{A}_T$ as a rich representation for spatial relations between activities of persons alongside the feature map $F$, the aggregation part aims to aggregate the information from all the persons to produce a final group activity label. The group activity is predicted by jointly processing the input features and the final activity map as:
\begin{equation}
\hat{p} = \zeta(F \oplus \hat{A}_T),
\end{equation}
where $\hat{p} \in R^{N_G}$ is the predicted probability vector of group activity classes. $\zeta$ uses convolutional and pooling layers on the concatenation of feature map and activity map to produce the final group activity predictions. If we denote a pooling layer with kernel size $x$ and stride 2 as $pool(x)$, $\zeta$ is composed of the sequence of layers $conv(7)$, $pool(2)$, $conv(7)$, $pool(2)$, $conv(7)$, $pool(2)$, $conv(1)$ followed by a global average pooling layer. This way the model will be able to globally extract the cues and predict the group activity. It is important for $\zeta$ to reason over the feature map and activity map jointly. If the activity map was the only input to the $\zeta$, it wouldn't be able to consider the matching between visual representations and corresponding activity map for the goal of recognizing the group activity. 

\subsubsection{Training}
CRM has a multi-task objective. It has to learn to predict the activity map $A_t$ for stage $t$ as accurate as possible. Moreover, it needs to produce the correct group activity label. 
Having the ground truth activity map $A$ for the individual and group activities, the loss functions are defined as follows;
\begin{equation}
  \mathcal{L} = w_A \mathcal{L_A} + w_G \mathcal{L}_G,
\end{equation}
\begin{equation}
  \mathcal{L}_{G}  = -\frac{1}{N_G}\sum_{i=1}^{N_G} \hat{p}_{i} log\left({p}_{i}\right), 
\end{equation}
\begin{equation}
  \mathcal{L_A} = \mathcal{L}_{A_1} + \mathcal{L}_{A_t} + \dots + \mathcal{L}_{A_T},
\end{equation}
\begin{equation}
    \mathcal{L}_{A_t} = \sum_{h=1}^{H^{'}} \sum_{w=1}^{W^{'}} \sum_{n=1}^{N} {(A_{t}^{h,w,n} - \hat{A}_{t}^{h,w,n})}^2,
\end{equation}
where $\mathcal{L_A}$ is the total activity map loss, $\mathcal{L}_{A_t}$ is the loss for the activity map of stage $t$, $\mathcal{L_G}$ is the group activity loss and, $\mathcal{L}$ is the total multi-task loss of the model. 
Also, $p$ denotes the one-hot-encoded group activity label.
$w_A$ and $w_G$ control the importance of total activity map and group activity losses, respectively. 

To make the model's job in training the multi-task loss easier, a two-step training strategy is used. In the first step, the $w_G$ is set to $0$. Therefore, the model concentrates on learning to produce the true activity map. After the first step, $w_G$ and $w_A$ are both given nonzero values to let all the weights optimize together. As the main goal is to predict the group activity, $w_A$ should be set smaller than $w_G$ so that more emphasis goes for the accurate group activity class prediction.

\section{Experiments}
In this section, we present experiments on two of the most important group activity datasets: Volleyball \cite{ibrahim2016hierarchical} and Collective Activity \cite{choi2009they}. Comparisons with baselines and state-of-the-art group activity recognition methods are provided to show the role of different components of the model and its superiority over the existing models.  

\subsection{Datasets}
\textbf{Volleyball Dataset}. In this dataset, there are 55 videos of volleyball matches. A number of short length clips are labeled in each video. 39 videos are used for training and the other 16 videos are used for testing. All the clips are consisted of 41 frames where only the middle frame is labeled with the bounding boxes and actions of persons along with the group activity label. Possible individual action labels are \textit{spiking}, \textit{blocking}, \textit{setting}, \textit{jumping}, \textit{digging}, \textit{standing}, \textit{falling}, \textit{waiting}, and \textit{moving}. Group activity labels are \textit{right spike}, \textit{left spike}, \textit{right set}, \textit{left set}, \textit{right pass}, \textit{left pass}, \textit{right winpoint}, and \textit{left winpoint}. 


\textbf{Collective Activity Dataset}. This dataset consists of 44 videos with different number of frames in each video.
The number of videos selected for train and test set are 31 and 13, respectively. Every 10th frame of all the videos is labeled with the bounding boxes of persons and their individual actions. The most frequent individual action label is considered as the group activity label in the labeled frame. The possible individual action and group activity labels are \textit{talking}, \textit{queuing}, \textit{waiting}, \textit{crossing}, and \textit{walking}. 

\subsection{Baselines}
We conduct experiments on various baselines to show the impact of each part of our model. Here, the reported results are on Volleyball dataset. The evaluated baselines are as follows.

\begin{enumerate}
   \item \textbf{Feature-Map-Only:} Only the feature map is fed into the aggregation component ($\zeta$). 
   \item \textbf{Activity-Map-Only:} $\zeta$ sees the stage 4 activity map without having access to the feature map.
    \item \textbf{Stage1-Activity-Map:} In this baseline, the initial activity map extracted by $\phi$ without further refinement is concatenated with the feature map to form the input of $\zeta$.
      \item \textbf{Stage2-Activity-Map:} The concatenation of stage 2 activity map and the feature map is used as the input of $\zeta$.
   \item \textbf{Stage3-Activity-Map:} The same as the previous baselines, except that the concatenation of stage 3 activity map and the feature map is used.
   \item \textbf{Stage4-Activity-Map:} Our complete model for single frame case which is similar to previous three baselines, but with refined activity map from the output of stage 4.
   \item \textbf{Stage4-Group-Activity-Map:} Similar to the previous baselines without the maps for individual activities.
   \item \textbf{Stage4-Activity-Map-Pool:}  In this baseline the aggregation component is replaced with a simple pooling mechanism in which the group activity with the highest sum over the boxes of the persons in the group activity fields is selected as the final group activity label. 
   \item \textbf{Stage4-Activity-Map-I3D-RGB:} In this baseline, RGB video frames are given to an I3D CNN backbone. Also, stage 4 refined activity map is used as input to $\zeta$. 

   \item \textbf{Stage4-Activity-Map-I3D-Flow:} This baseline is similar to the previous baseline except that a stack of optical flow fields between the frames is the input of I3D. 
   \item \textbf{Stage4-Activity-Map-I3D-Fusion:} This is our final model for multi-frame input which is the average fusion of the results of two previous baselines. 

\end{enumerate}

\subsection{Implementation Details}
We use Tensorflow \cite{abadi2016tensorflow} to implement our model. Adam optimizer \cite{kingma2014adam} with different learning rates is used to train our models. Inception-V3 \cite{szegedy2016rethinking} and I3D \cite{carreira2017quo} are the backbone CNN architectures used for the single and multi-frame cases, respectively. To extract the feature map using the Inception-V3, the output of the \textit{Mixed\_7c} layer is utilized. The \textit{Mixed\_4f} layer is the feature map extraction layer for the I3D. The extracted feature maps of Inception-V3 and I3D are resized to form a $H^{'} \times W^{'} \times D$ feature map. Additional to RGB stream of the I3D, we use stacked optical flow fields as input to another stream of I3D in the multi-frame setting and combine the results of group activity probabilities with the RGB I3D stream using simple averaging of the predictions. We use TVL1 algorithm \cite{zach2007duality} to compute these optical flow fields. In the multi-frame scenario, we use the middle frame, 5 frames before it and 4 frames after it as the input to the model in both datasets.

\textbf{Volleyball Dataset.} Input frames are resized to $720\times1280$. We also consider the size of $43\times78$ for the activity map and therefore resize the feature map to this size. In the first step of the training, we train the model using $w_G=0$ and $w_A=1$ for 10 epochs with the learning rate of 0.00001 and another 10 epochs with the decreased learning rate of 0.000001. In the joint training step, the $w_G=1$ and $w_A=0.0001$ so that the model concentrates on the group activity recognition. This time the model is trained with learning rate 0.00005 for 10 epochs then it is trained for another 10 epochs with the learning rate of 0.000005.

\textbf{Collective Activity Dataset.} Resized images are $240\times360$ for the RGB inputs and $480\times720$ for the optical flow fields. The resized feature maps are $30\times45$. For both modalities, the first step of the training (the first 20 epochs) is the same as the training procedure for Volleyball dataset described above. In the second step, the model for optical flow modality is trained for two consecutive 10 epochs with the learning rates of 0.00001 and 0.000001, respectively. The Second step of training for the RGB modality consists of 3 and 2 epochs of training with the learning rates of 0.00005 and 0.000005, respectively. In the training procedure of models for both modalities, we set $w_G=1$ and $w_A=0.001$ in the second step.

\begin{table}
\begin{center}
\begin{tabular}{|l|c|}
\hline
Method & Accuracy \\
\hline\hline
Feature-Map-Only & 75.99 \\
Activity-Map-Only & 82.72 \\
Stage1-Activity-Map & 89.82 \\
Stage2-Activity-Map & 90.72 \\
Stage3-Activity-Map & 90.42 \\
Stage4-Activity-Map & 90.80 \\
Stage4-Group-Activity-Map & 88.85 \\
Stage4-Activity-Map-Pool & 87.80\\
\hline
Stage4-Activity-Map-I3D-RGB & 92.07 \\
Stage4-Activity-Map-I3D-Flow & 91.47 \\
Stage4-Activity-Map-I3D-Fusion & \textbf{93.04} \\
\hline
\end{tabular}
\end{center}
\caption{Various baselines and our final models in the single frame and multiple frame settings. The volleyball dataset is used in these experiments.}
\label{table:baselines}
\end{table}

\subsection{Analysis}
To fully understand the impact of every module of our proposed model, different variants of this model listed above as the baselines are applied to the Volleyball dataset. The results of all baselines are reported in Table~\ref{table:baselines}. As expected, Feature-Map-Only baseline has the lowest accuracy because it does not have access to additional information of activity map in any form. The Activity-Map-Only baseline with access to only the activity map performs better than the Feature-Map-Only showing that the activity map is a richer representation than the feature map for group activity recognition. However, as shown in the following a combination of both maps provides much better results. 

To consider the joint use of feature and activity map, we compare the effect of their combination in two settings. First, we observe that using the stage 1 activity map produced by $\phi$ without further refinement boosts the performance of the model from 75.99\% in the Feature-Map-Only model to 89.82\%. This shows that even without the refinement stages, the presence of the activity map greatly improves the performance of the proposed model in recognizing the group activity. This is due to the fact that in our model $\zeta$ learns to aggregate the visual and temporal information from the feature map with the spatial relations between the activities of persons presented by the activity map.
Second, the stage 1 activity map is refined in multiple consecutive stages.
The refinement stages ($\psi_t$) will correct the predictions of $\phi$ by applying additional convolutional layers whose effective receptive fields increase with $t$. This enables the model to consider the spatial information in a larger neighborhood and provide more accurate activity maps. As the results of Table~\ref{table:baselines}  show, for the case of Volleyball dataset, the performance of Stage2-Activity-Map has an improvement of 1\% compared to Stage1-Activity-Map. The performance of Stage3-Activity-Map and Stage4-Activity-Map models are at the same level as that of Stage2-Activity-Map. Although we use 4 stages, it is also possible to consider only two stages with the cost of about 0.1\% accuracy in the cases where the computational efficiency is important.

Labeling individual actions for every person in a scene is a time consuming task and it may not be available in every dataset. Therefore, in the Stage4-Group-Activity-Map baseline, we evaluate the performance of our model by constructing the activity map using only the group activity fields. Achieving an accuracy of 88.85\% shows that the activity map representation still offers valuable information even without the individual activities. However, the inclusion of individual activities in Stage4-Activity-Map model shows 2\% improvement of accuracy compared to Stage4-Group-Activity-Map. 

It is possible to extract group activities from the activity map without using the aggregation component $\zeta$. In the Stage4-Activity-Map-Pool baseline, to infer the group activity label, sum of the values inside the location of bounding boxes of all persons are calculated for all the group activity fields inside the activity map. This results in scores for each group activity class and the class with the highest score is chosen as the final prediction. The accuracy of this baseline is 87.80\% which is lower than the complete model with the aggregation part $\zeta$. Without the $\zeta$ model is not able to do a global aggregation on the scene features and the activities that are happening in it. Therefore, it is necessary to have another reasoning component over the activity map to be able to make robust predictions. It is possible to extract individual activities from the activity map in a similar approach to Stage4-Activity-Map-Pool. After inferring individual activities in this way, an accuracy of 78.59\% is achieved. Better performance can be accomplished by introducing individual aggregation components for activities of persons but it is not the focus of our work. 

Temporal information is very important in activity recognition. The I3D CNN is able to extract some of this information. Changing the backbone to I3D CNN with multiple frames as input in Stage4-Activity-Map-I3D-RGB leads to a better model with 92.07\% accuracy. 
The motion feature is a special kind of temporal information. 
3D CNNs can also work on stacked optical flow fields to extract motion specific information. Stage4-Activity-Map-I3D-Flow is another baseline model in which I3D is applied to the optical flow fields instead of the RGB frames leading to an accuracy of 91.47\%.
Predictions of stacked RGB and optical flow models can be fused to make a stronger model (Stage4-Activity-Map-I3D-Fusion). Here, we simply take the average of predicted probabilities by the models to produce the final probability vectors. The accuracy of the fused model is 93.04\% which shows the positive effect of the fusion step. 

\begin{figure}[t]
\begin{center}
   \includegraphics[trim=0 0 0 0, width=1.0\linewidth]{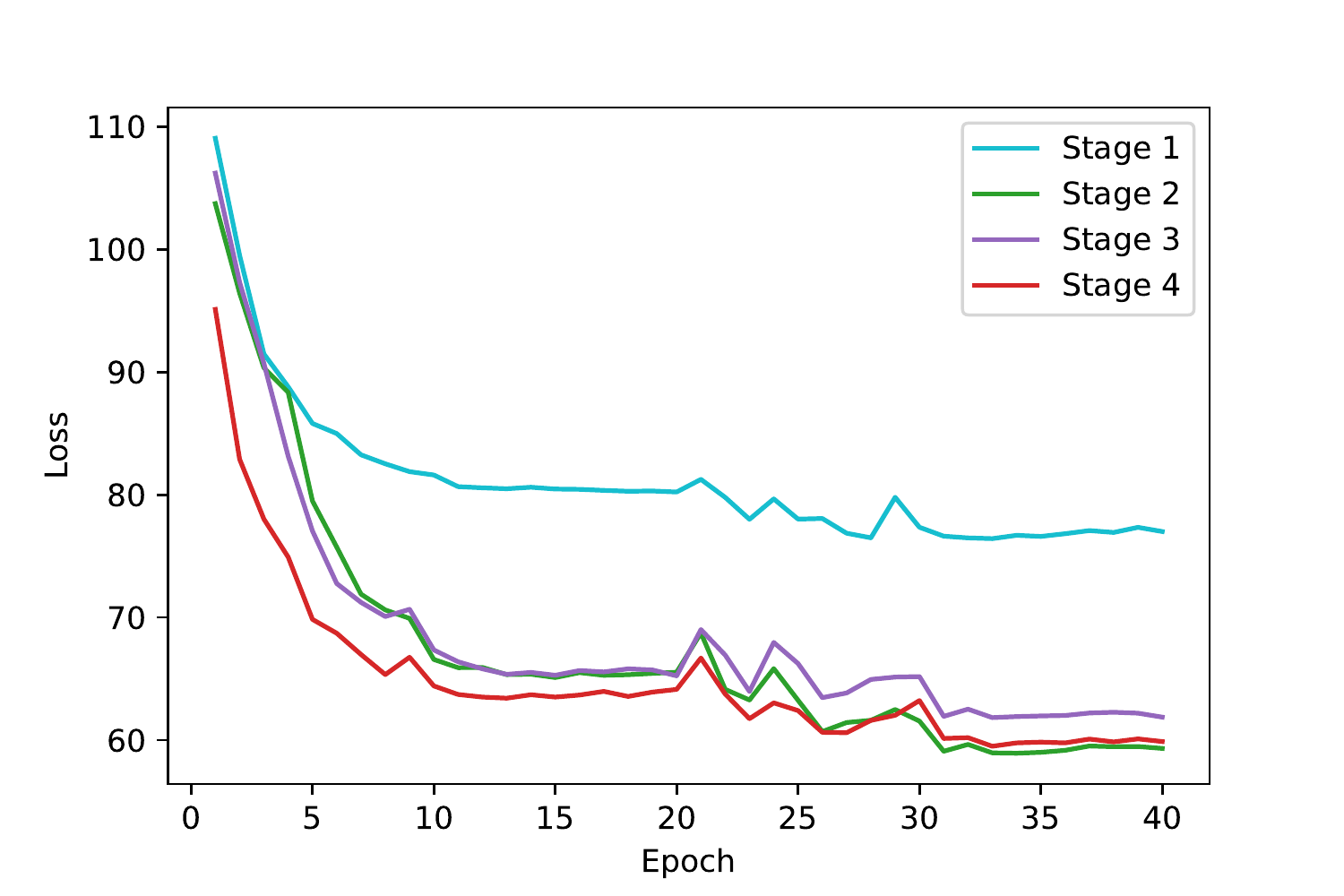}
\end{center}
   \caption{Activity map losses of stages 1 to 4 during the two-step training procedure for the Stage4-Activity-Map-I3D-RGB baseline.}
\label{fig:loss}
\end{figure}

To analyze the impact of the refinement stages, the losses of different stages on the test data during different epochs of training for the I3D with RGB inputs are shown in the Figure~\ref{fig:loss}. During the first step of the training where the model is only concentrated on the task of minimizing the activity map loss, each stage decreases the loss and makes a better activity map. However, in the second step of the training starting from the 21st epoch, the losses of all stages are better than the first stage activity map but due to the difficulties of the model in minimizing two losses related to activity map and the group activity, the behaviour of losses become less stable. However, it is guaranteed that the stage 2, 3 and 4 losses are better than stage 1 but there may be small unexpected differences between the losses of refinement stages.  The small value of $w_A$ is one of the reasons for this minor problem because less importance is given to the accurate activity map prediction and the focus is on predicting the group activity. Therefore, the gradients of $\zeta$ can be harmful to the task of activity map prediction. The gradient flow from $\zeta$ to the refinement stages can have larger impacts on the final stages. This effect is slightly reduced by reaching the earlier stages. This explains why the losses in the middle stages got closer to the stage 4 loss. This problem is inevitable and it can happen with less or more stages. Here, the ultimate goal is to predict group activity labels as accurately as possible and after the first stage, the small inconsistencies can be ignored. 

\begin{figure*}[t]
\begin{center}
   \includegraphics[trim=0 0 0 0, width=1.0\linewidth]{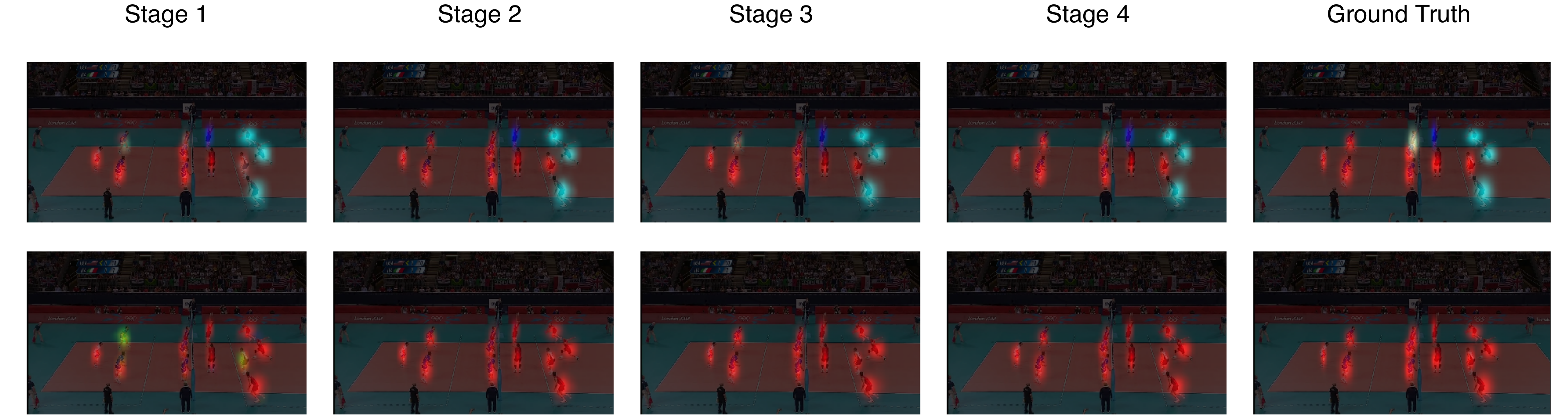}
\end{center}
   \caption{Samples of the generated activity map in different refinement stages. Top row includes the combined individual activity map with different colors for each individual activity class (9 distinct colors). Also, the separated group activity map is visualized in the bottom row considering different colors for the group activity classes. Similar colors in the top and bottom row do not represent the same class. Best viewed in color.}
\label{fig:sample}
\end{figure*}

A visualization of the generated activity map in different stages is provided in Figure~\ref{fig:sample}. There are noticeable errors in the first stage of both individual and group activity fields in the activity map. Starting from the second stage, the effect of the refinement stages is visible. For example, the model in the first stage considers wrong group activities for two persons but the refinement stages are able to fix these errors after observing the predictions made for the other neighbouring persons. Both the individual and group activity fields are visible to the refinement stages in a specific local area based on the receptive fields of the layers which helps them refine the predictions based on the other local predictions. 

\subsection{Comparison to the State of the Art}
\textbf{Volleyball Dataset. } The results of our model in both multi-frame and single frame cases are compared with other methods in the Table~\ref{table:state_of_the_art}. Not only does the single frame CRM outperform previous approaches in the single frame case by a large margin but also achieves better results than the temporal versions of those approaches.
In the single frame scenario, CRM is about 2.5\% better than \cite{ibrahim2018hierarchical}. Although their model includes the relational information between the individuals, it doesn't use all the information in spatial relations including the spatial location of individuals relative to each other. Therefore, our model is able to outperform it. 
Considering the accuracy of 93.04\% for the temporal version of our model, a significant improvement of about 2.4\% is achieved compared to the previous best performing model of \cite{bagautdinov2017social} (about 26\% of possible improvement to achieve 100\% accuracy). In \cite{bagautdinov2017social}, the group activity representation is generated by a simple pooling strategy disregarding most of the relational cues. However, our model tries to extract all the relational information in the scene.

\textbf{Collective Activity Dataset. } For Collective Activity dataset, group activity performance of the CRM is evaluated in two different settings. 
First, the Multi-class Classification Accuracy (MCA) of our model on the Collective Activity dataset is compared with other approaches in the Table~\ref{table:collective}.
As the reported results of this table show, the proposed model has competitive results with the best performing methods of \cite{shu2017cern} and \cite{li2017sbgar} and outperforms other approaches.
Although the \cite{shu2017cern} is about 1.5\% better than our model in Collective Activity dataset, it falls far behind CRM on Volleyball dataset with about 10\%.  
\textit{Walking} and \textit{Crossing} activities in the Collective Activity dataset are simply the same activities performed at different locations. Therefore, similar to \cite{wang2017recurrent}, we combine their predictions into the new activity of \textit{Moving}. The Mean per Class Accuracy (MPCA) for the new setting is reported in Table~\ref{table:collective_moving}. 
The confusion matrix is needed for calculating the MPCA for the new 4 classes. Therefore, due to the lack of confusion matrix in \cite{shu2017cern}, we couldn't report their results in this part. 
According to the results, CRM outperforms other approaches including \cite{li2017sbgar} in this setting considering the MPCA as the evaluation metric. It is due to the fact that most of its incorrect predictions were because of the natural confusion between \textit{Walking} and \textit{Crossing} activities. Therefore, it is evident that CRM is able to achieve notable performance in Collective Activity dataset. 

\begin{table}
\begin{center}
\begin{tabular}{|l|c|c|}
\hline
\multicolumn{1}{|l|}{{Method}} & {Multiple Frames} & {Single Frame} \\ \hline
HDTM \cite{ibrahim2016hierarchical}    & 81.90    & -            \\ \hline
CERN \cite{shu2017cern}     & 83.30    & -            \\ \hline
Social Scene \cite{bagautdinov2017social}   & 90.60     & 83.80         \\ \hline
HRN \cite{ibrahim2018hierarchical}   & 89.50     & 88.30         \\ \hline
CRM                         & \textbf{93.04}       & \textbf{90.80}      \\ \hline
\end{tabular}
\end{center}
\caption{Comparison of our results with those of the state-of-the-art methods in multiple or single frame cases.}
\label{table:state_of_the_art}
\end{table}

\begin{table}
\begin{center}
\begin{tabular}{|l|c|}
\hline
\hspace{0.1cm}Method\hspace{0.1cm}                   & \multicolumn{1}{l|}{\hspace{0.1cm}Accuracy\hspace{0.1cm}} \\ \hline
\hspace{0.1cm}\cite{choi2012unified}\hspace{0.1cm}         & 80.40                         \\ \hline
\hspace{0.1cm}\cite{hajimirsadeghi2015visual}\hspace{0.1cm}  & 83.40                         \\ \hline
\hspace{0.1cm}\cite{lan2012discriminative}\hspace{0.1cm}                 & 79.70                        \\ \hline
\hspace{0.1cm}HDTM \cite{ibrahim2016hierarchical}\hspace{0.1cm}        & 81.50                         \\ \hline
\hspace{0.1cm}SBGAR \cite{li2017sbgar}\hspace{0.1cm} & 86.10                       \\ \hline
\hspace{0.1cm}CERN \cite{shu2017cern}\hspace{0.1cm}                 & \textbf{87.20}                \\ \hline
\hspace{0.1cm}CRM-RGB\hspace{0.1cm}                &        83.41                  \\ \hline
\hspace{0.1cm}CRM-Flow\hspace{0.1cm}                &       85.44                   \\ \hline
\hspace{0.1cm}CRM\hspace{0.1cm}                &            85.75              \\ \hline
\end{tabular}
\end{center}
\caption{Comparison of the MCA of CRM with the other approaches on Collective Activity dataset.}
\label{table:collective}
\end{table}

\begin{table}[t]
\begin{center}
\begin{tabular}{|l|c|c|c|c|c|}
\hline
Method                   & {M}& {W}& {Q}& {T} & {MPCA}\\ \hline
\cite{choi2012unified}         & 90.0 & 82.9 & 95.4 & 94.9  & 90.8   \\ \hline
\cite{hajimirsadeghi2015visual}& 87.0 & 75.0 & 92.0 & 99.0  & 88.3   \\ \hline
\cite{lan2012discriminative}   & 92.0 & 69.0 & 76.0 & 99.0  & 84.0   \\ \hline
HDTM \cite{ibrahim2016hierarchical} & \textbf{95.9} & 66.4 & 96.8 & \textbf{99.5}  & 89.7   \\ \hline
SBGAR \cite{li2017sbgar}             & 90.8 & 81.4 & 99.2 & 84.6  & 89.0   \\ \hline
\cite{wang2017recurrent}       & 94.4 & 63.6 & \textbf{100.0} & \textbf{99.5}  & 89.4   \\ \hline
CRM                           & 91.7 & \textbf{86.3} & \textbf{100.0} & 98.91  & \textbf{94.2}  \\ \hline
\end{tabular}
\end{center}
\caption{The mean per class accuracies (MPCA) and per class accuracies of our model in comparison to other methods on Collective Activity dataset. M, W, Q, T are the abbreviations for Moving, Waiting, Queuing, and Talking, respectively.}
\label{table:collective_moving}
\end{table}


\section{Conclusions}
We propose a Convolutional Relational Machine  for group activity recognition by extracting the relationships between persons. We show that the activity map is a useful representation that effectively encodes the spatial relations. We also show that an aggregation method is necessary for the refined activity map to produce reliable group activity labels. Future work can adapt this model to extract the spatial relations in person-object scenarios. 

\newpage

{\small
\bibliographystyle{ieee_fullname}
\bibliography{egbib}
}

\end{document}